\documentclass[twoside,leqno,twocolumn]{article}

\usepackage[letterpaper]{geometry}

\usepackage{ltexpprt}
\usepackage{hyperref}
\usepackage{latexsym}
\usepackage{amssymb}
\usepackage{amsmath}
\usepackage{booktabs}
\usepackage{enumitem}
\usepackage{graphicx}
\usepackage{color}
\usepackage{pgfplots}
\pgfplotsset{compat=newest}
\usepackage{mathtools}
\usepackage{algorithm}
\usepackage{algorithmic}
\usepackage{multirow}
\usepackage{biblatex}
\begin{document}

\newcommand\relatedversion{}

\title{\Large Graph Neural Networks as Ordering Heuristics for Parallel Graph Coloring}

\author{Kenneth Langedal \and Fredrik Manne}

\date{University of Bergen}

\maketitle







\begin{abstract} 
The graph coloring problem asks for an assignment of the minimum number of distinct colors to vertices in an undirected graph with the constraint that no pair of adjacent vertices share the same color. The problem is a thoroughly studied NP-hard combinatorial problem with several real-world applications. As such, a number of greedy heuristics have been suggested that strike a good balance between coloring quality, execution time, and also parallel scalability. In this work, we introduce a graph neural network (GNN) based ordering heuristic and demonstrate that it outperforms existing greedy ordering heuristics both on quality and performance. Previous results have demonstrated that GNNs can produce high-quality colorings but at the expense of excessive running time. The current paper is the first that brings the execution time down to compete with existing greedy heuristics. Our GNN model is trained using both supervised and unsupervised techniques. The experimental results show that a 2-layer GNN model can achieve execution times between the largest degree first (LF) and smallest degree last (SL) ordering heuristics while outperforming both on coloring quality. Increasing the number of layers improves the coloring quality further, and it is only at four layers that SL becomes faster than the GNN. Finally, our GNN-based coloring heuristic achieves superior scaling in the parallel setting compared to both SL and LF.
\end{abstract}

\section{Introduction} 

Given an undirected graph $G=(V, E)$ where $V$ is the set of vertices and $E$ is the set of edges. A graph coloring of $G$ is an assignment of colors $c:V\rightarrow\mathbb{Z}$ to each vertex $u$ of the graph such that no adjacent vertices share the same color, i.e., $\forall \{u, v\} \in E \implies c(u) \not = c(v)$. The smallest number of distinct colors needed to color a graph is known as its chromatic number, and the decision version of the problem was one of Karp's original NP-complete problems \cite{karp2010reducibility}. Graph coloring has several real-world applications. Examples of this include scheduling \cite{arkin1987scheduling}, air traffic flow management \cite{barnier2004graph}, and designing timetables \cite{lewis2015timetable}. For an overview of the problem's history and early results, see Voloshin \cite{voloshin2009graph}.

Numerous algorithms have been developed for the graph coloring problem, including exact algorithms and heuristics. The heuristics can be further divided into those that search for improvements, such as \textit{local search} based approaches \cite{galinier2006survey}, and \textit{greedy heuristics} that consider the vertices in a specific order and colors the vertices greedily using the smallest available color \cite{welsh1967upper}. This work focuses exclusively on greedy heuristics for shared memory multicore computers. These greedy heuristics are mainly motivated by their low running time while at the same time producing solutions with a small numbers of colors. In numerical computations such as solving PDEs and algorithmic differentiation it is often required to repeatedly perform graph colorings to partition a set of tasks into independent sets. Since the coloring is only used to set up the ensuing numerical computation, it is not feasible to spend substantial amounts of time on it, and it is also only required to have a good, but not optimal, solution. In this setting, fast greedy heuristics are often the only choice.

We show that shallow graph neural networks with few parameters can compete and even 
outperform existing ordering heuristics on solution quality, execution time, and parallel scalability. Our main results are as follows:
\begin{itemize}
    \item Small models with only two layers can run faster than the simplest non-trivial ordering heuristic SL. While producing higher-quality colorings than both SL and LF.
    \item Larger models with more layers are able to produce orderings that use less colors, and only at four layers does the GNN-based heuristic become slower than SL.
    \item On a multicore computer our GNN model scales better than corresponding parallel greedy coloring heuristics.
    \item Experimental results on large graphs, including one instance with over one billion edges validates the performance of the GNN model.
\end{itemize}

\section{Related Work}

The first greedy heuristic for graph coloring was introduced by Welsh and Powell \cite{welsh1967upper}. This colors each vertex according to a predetermined ordering with the smallest available color. It follows that the resulting coloring uses at most $\Delta + 1$ colors, where $\Delta$ is the largest degree in the graph. This same greedy approach can be used with any ordering of the vertices. In increasing order of execution time and coloring quality, these are the most commonly used ordering rules for greedy coloring:
\begin{itemize}
    \item \textbf{First Fit (FF)}. Color the vertices in the order they appear in the input \cite{welsh1967upper, lovasz1989line}.
    \item \textbf{Largest Degree First (LF)}. Color the vertices in decreasing order of degree \cite{welsh1967upper}.
    \item \textbf{Smallest Degree Last (SL)}. Color the vertices in the reverse order induced by iteratively selecting and removing the smallest degree vertex from the graph \cite{matula1983smallest}.
    \item \textbf{Saturation Degree (SD)}. The saturation degree of a vertex is the number of distinct colors used by its colored neighborhood. The saturation degree heuristic colors the vertices by iteratively coloring the vertex with the highest saturation degree \cite{brelaz1979new}.
\end{itemize}

The intuition for LF, SL and SD is that each vertex is ordered by how difficult it will be to assign a low numbered color to it. Thus one tries to first color the vertices most likely to give rise to a new color class. All of these ordering heuristics followed by greedy coloring can be made to run in $\mathcal{O}(V + E)$ time \cite{hasenplaugh2014ordering, matula1983smallest}.

In addition to these, there are also \textit{log-based} versions of the LF and SL heuristics by Hasenplaugh et al.~\cite{hasenplaugh2014ordering}. The log-versions work similarly, except they use the log-degree rounded-up to order the vertices. These heuristics generally perform equal or slightly worse than the normal LF and SL heuristics but offer better guarantees for parallel scalability. In the parallel version of SL, every vertex of smallest degree is removed simultaneously, thus deviating slightly from the sequential version \cite{allwright1995comparison, besta2020high, hasenplaugh2014ordering}. Hasenplaugh et al.~report that this formulation also results in fewer colors used.

Incidence degree (ID) orders the vertices by decreasing number of already colored neighbors \cite{newsam1983estimation}. We mainly skip it in our study as it does not have a good parallel implementation and generally produces colorings of roughly the same quality as SL. The SD heuristic is included for its superior coloring quality, even though it does not yet have any good parallel implementation.

The second part of an ordering heuristic is the actual coloring of the vertices following the produced ordering. Sequentially, coloring the vertices in the given order is straightforward while always using the smallest available color. This approach will be referred to as just \textit{greedy}. As already stated, this will never use more than $\Delta+1$ colors, where $\Delta$ is the maximum vertex degree. Also note that there always exists some ordering such that greedy will produce a coloring using the minimum number of colors, i.e.~it uses the same number of colors as the chromatic 
number. For more information on greedy coloring algorithms and their use see 
Gebremedhin et al.~\cite{gebremedhin2013colpack}.

For parallel coloring, the Jones-Plassmann (JP) algorithm \cite{jones1993parallel}
gives a framework that can exploit orderings while still coloring several vertices concurrently. The general idea behind the JP algorithm is to direct the edges of $G$ such that it becomes a DAG (Directed Acyclic Graph). Then starting with the sources, i.e.~the vertices with in-degree 0, one can color any set of vertices concurrently as long as all their predecessors in $G$ have already been colored. JP maintains the $\mathcal{O}(V+E)$ running time, with expected parallel running time of $\mathcal{O}(\text{log}(V)/\text{log log}(V))$ under the PRAM model. However, it is worth noting that for adversarial inputs, JP can result in $\Omega(V)$ span, meaning there is no room for parallel work.

To run the sequential greedy coloring heuristic one does not need to have a total ordering of the vertices, it is sufficient to determine for each edge $(u,v)$ which one of $u$ and $v$ should be colored first. This defines a partial ordering on the vertices that can be used to create the necessary DAG for the JP algorithm. This is done by assigning each vertex a \textit{priority} $p : V \rightarrow \mathbb{R}^+$ that describes the position of a vertex in the ordering, thus if $p(u)>p(v)$, then $u$ comes before $v$ in the ordering. 

The outline of the JP algorithm can be seen in Algorithm \ref{alg:JP}. The first parallel for-loop partitions the neighbors of a vertex $u$ into those that should be colored before and after $u$. It also computes the set of all sources in $G$ that
can be colored initially. The second parallel for-loop colors the current set of vertices where all predecessors have already been colored. In doing so it also checks if any successors to this set are ready to be colored in the next round. This loop runs until all vertices have been colored.

For both greedy and JP it is sufficient to only generate a partial ordering on the vertices. This significantly reduces the work required without affecting the performance of the parallel coloring using JP. It also means that any parallel procedure for generating an ordering can be combined with JP to get a complete parallel coloring algorithm. This combination will result in the same number of colors as the sequential counterpart. The initial ordering could come from some ordering heuristic or from a GNN.

\begin{algorithm}[H]
\begin{algorithmic}[1]
\REQUIRE $G=(V,E,p)$
\ENSURE Assignment of colors $c:V\rightarrow\mathbb{Z}$
\STATE $next \gets \emptyset, prev \gets \emptyset$
\STATE \textbf{parallel for} $u \in V$
\STATE \hspace{\algorithmicindent} $pred(u) \gets {v \in N(u) : p(v) > p(u)}$
\STATE \hspace{\algorithmicindent} $succ(u) \gets {v \in N(u) : p(u) > p(v)}$
\STATE \hspace{\algorithmicindent} $count(u) \gets |pred(u)|$
\STATE \hspace{\algorithmicindent} \textbf{if} $count(u) = 0$
\STATE \hspace{2\algorithmicindent} \textbf{atomic} $prev \gets prev + \{u\}$
\STATE
\STATE \textbf{while } $prev \not = \emptyset$
\STATE \hspace{\algorithmicindent} \textbf{parallel for} $u \in prev$
\STATE \hspace{2\algorithmicindent} $c(u) \gets$ min color among $pred(u)$
\STATE \hspace{2\algorithmicindent} \textbf{for} $v \in succ(u)$
\STATE \hspace{3\algorithmicindent} \textbf{atomic} $count(v) \gets count(v) - 1$
\STATE \hspace{3\algorithmicindent} \textbf{if} $count(v) = 0$
\STATE \hspace{4\algorithmicindent} \textbf{atomic} $next \gets next + \{v\}$
\STATE \hspace{1\algorithmicindent} $swap(next,prev)$

\end{algorithmic}
\caption{Pseudocode for the JP parallel coloring algorithm as used in this work. Implemented using local buffers for \textit{next} and \textit{prev}, and atomic \textit{sub-fetch} operations for \textit{count}.}
\label{alg:JP}
\end{algorithm}

To the best of our knowledge, no attempts have been made to use machine learning to compete directly with greedy graph coloring heuristics on execution time, parallel scalability, and solution quality. Huang et al.~\cite{huang2019coloring} compared their AlphaGoZero-based approach against greedy heuristics on solution quality, but the approach was orders of magnitude slower. In the broader view of graph coloring, there are several examples where machine learning was used \cite{ijaz2022solving, lemos2019graph, li2022rethinking}. Common amongst these is that they consider the $k$-coloring version of graph coloring. This is a decision problem that asks if a graph can be colored with at most $k$ colors for some given value $k$. The training can then be formulated as minimizing the number of adjacent vertices sharing the same color. For example, Schuetz et al.~\cite{schuetz2022graph} trained GNN models to output colors directly using a physics-inspired framework. The high-level approach was to train one model per instance, where an instance is a graph and $k$. Then, use random initial inputs to allow equal neighbors to receive different colors. The size and execution times are very different from the setting with greedy heuristics, using hours on graphs with only a few hundred vertices. For reference, SL uses less than a second on graphs with millions of vertices.

\section{GNN Ordering Heuristic}

The proposed approach uses graph neural networks to directly output priority values for each vertex and then greedily color the graph using either greedy for sequential coloring or JP for parallel coloring. The overall strategy is to perform supervised learning on the SL and SD orderings and then leverage the ease of parallelism offered by GNNs to compute the approximate ordering faster. The rest of this section outlines the decisions made regarding GNN architecture and the training process.

\subsection{GNN Architecture}

There is a rapidly growing number of GNN architectures available for use. At the time of writing, PyTorch Geometric provides more than 40 GNN architectures \cite{PyG}. It has already been demonstrated that GNNs can find high-quality colorings that outperform greedy heuristics \cite{huang2019coloring}. However, in this work, we also aim to outperform greedy heuristics in terms of time and scalability. Since the greedy heuristics we compete against are fast and simple, only the simplest GNN architectures are relevant. Among these, the graph convolutional networks (GCN) \cite{kipf2016semi} and graph sample and aggregate (GraphSAGE) \cite{hamilton2017inductive} architectures stand out as good candidates. For $k$-coloring, Schuetz et al.~\cite{schuetz2022graph} already found GraphSAGE to perform better than GCN. Although the scale and problem are different, it provides some reassurance moving forward with GraphSAGE.

\subsection{GraphSAGE}

The GNN model consists of several \textit{layers} stacked on top of each other. A graph $G=(V, E)$ is given as input to the model, where each vertex has a feature representation that changes after every layer in the model. At layer $l$, the feature representation for $u\in V$ is stored in a vector $H^{(l)}_u$. The length $d$ of the feature vector at layer $l$ is denoted by $d^{(l)}$. Stacking all the feature vectors at the $l$'th layer gives the matrix $H^{(l)} \in \mathbb{R} ^{|V| \times d^{(l)}}$. Independent from any input graph, every layer in the GNN model has trainable parameters $W^{(l)} \in \mathbb{R} ^{2d^{(l)} \times d^{(l+1)}}$, bias $b^{(l)} \in \mathbb{R}^{1 \times d^{(l+1)}}$, and a non-linear activation function $\sigma$, such as $\textrm{ReLU}(x)\coloneqq\max(0,x)$. With this, we use the following layer-wise propagation rule.

\begin{algorithm}
\begin{algorithmic}[1]
\caption{GraphSAGE propagation rule. $T_u$ is a temporary variable holding the aggregated feature vectors of the neighbours of $u$.}
\label{alg:SAGE}
\REQUIRE $H^{(l)}$
\ENSURE $H^{(l+1)}$
\STATE \textbf{for} {$u \in V$}
\STATE \hspace{\algorithmicindent} $T_u \gets \sum_{v \in N(u)} H^{(l)}_v / |N(u)|$
\STATE \hspace{\algorithmicindent} $H^{(l+1)}_u \gets \sigma(W^{(l)}\cdot \text{CONCAT}(H^{(l)}_u, T_u) + b^{(l)})$
\end{algorithmic}
\end{algorithm}

This propagation rule is not identical to the one provided in \cite{hamilton2017inductive}. However, it retains the main distinguishing feature of GraphSAGE compared to GCN, namely the concatenation of a vertex's own features before the transformation step.

In the case of graph coloring, there are no features associated with the vertices in the input. Therefore, we set the input dimension $d^{(0)}=2$ and initialize $H^{(0)}$ with the degree and ID for each vertex in the graph. The reason for including the ID is to let the model learn the same tie-breaking rule that will be used to generate training data. While this breaks the permutation invariance of the model, the following coloring by JP or greedy would not have this property anyway. At some point, there would need to be a tie-breaking mechanism. Beyond degree and ID, it would be possible to do feature engineering and extract further input features, such as neighborhood connectivity or some centrality measure. However, it is already a challenge to bring the running time down to a level where it can compete with heuristics such as SL. Therefore, we do not perform any additional feature engineering.

To train a model, we propose a two-stage procedure consisting of:
\begin{itemize}
    \item Supervised learning using vertex orderings from other greedy coloring heuristics as training data.
    \item Population-based genetic training that directly mutates the parameters of the models to improve the number of colors used.
\end{itemize}

\subsection{Supervised Training}

The first challenge is finding a good representation of a vertex ordering for supervised learning. The order produced by any of the previously mentioned ordering heuristics is an assignment of priority values for each vertex. It would be possible to train directly on these priority values. However, there could be a large number of distinct priorities, making it difficult to formulate it as a classification task. A nice property of the JP algorithm is that it only needs to compare priorities for neighboring vertices. Therefore, supervised training can be formulated as a binary classification task on the edges of the graph. Since the order of the endpoints matters, each undirected edge $(u,v) \in E$ is considered as two directed edges $(u,v)$ and $(v,u)$. Then, each directed edge $e=(u,v)$ is classified as follows.
\begin{equation*}
  C(e) =
    \begin{cases}
      1 & \text{if $P(u) > P(v)$}\\
      0 & \text{otherwise}
    \end{cases}
\end{equation*}

One potential problem with using edge classification is that the output from the model might not be a partial ordering over the vertices. This means it could contain cycles, which would cause issues for JP. To solve this, we only use edge classification during training and use the last node embedding as the final output at inference. More specifically, let $H_u$ be the feature vector of vertex $u$ after the last GraphSAGE layer of the model. Then, at inference, we set the priority $P(u) = sum(H_u)$. This avoids any cycles as long as ties are handled consistently. To get edge labels during the training stage, the label for each edge $e=(u,v)$ is defined as $\sigma(sum(H_u) - sum(H_v))$. This way, if the model wants to classify the edge as 1, meaning $u$ should come before $v$, it needs to produce node embeddings such that $sum(H_u) > sum(H_v)$, which is exactly what we want.

For the more complicated ordering heuristics, it is necessary to handle ties consistently. For instance, in the SD heuristic, any uncolored vertex with the largest saturation degree could be chosen. When there is a tie, the vertex with the highest degree is preferred, and when there is still a tie, the vertex with the lowest ID is selected. By providing these features as input to the model, learning even the more complicated heuristics remains plausible.

Ideally, a model that learns to imitate a better heuristic like SD with significant accuracy should also produce orderings that use roughly the same number of colors. However, there is no guarantee that this will be the case. Since the overall goal for a trained model is to produce orderings that use as few colors as possible, we also employ a second genetic-based training stage. This stage keeps improving the model parameters found during the supervised learning stage. Genetic-based training has seen success in certain areas where reinforcement learning would typically be used, such as playing Atari games~\cite{such2017deep, faycal2022direct}.

\subsection{Genetic Training}

Our training procedure is similar to that presented by Such et al.~\cite{such2017deep} and works as follows.

It starts from a population $P$ consisting of $N$ graph neural networks, identified by their parameter vector. Each generation undergoes a selection and mutation stage. Truncation selection is used, where the top $T$ performing networks are used as parents for the next population. To build up a network for the next generation, a new network starts from one or two parent networks picked uniformly at random. Then, a random mutation or crossover operator is selected to produce the new network. Finally, the top performer is brought over to the next generation unchanged.

The performance of a network is defined as the number of colors used to color the entire training dataset. Since there can be plateaus where any small change in parameters does not change the number of colors used, the number of vertices colored with the highest color serves as a tiebreak. For example, if two orderings use 10 colors, but one uses color 10 five times and the other three times, the second order is considered better.

For genetic operators, we use a combination of operators from the literature designed specifically for neural networks. The simplest operator mutates the parameters of a single parent by applying random Gaussian noise, as used by Such et al.~\cite{such2017deep}. Monatana et al.~\cite{montana1989training} introduced a slightly more complicated operator that takes the network's internal structure into account. Instead of applying random noise to all the parameters, it selects a subset of the parameters that contribute to the same internal features and only mutates those. Finally, we use a more recent addition to genetic operators for neural networks by Faycal and Zito \cite{faycal2022direct}. This operator only uses the most significant parameters from each parent, meaning the parameters with the highest absolute value, and then fills in the remaining values with new random values. During training, all the operators mentioned above are used.

Overall, our main emphasis is on supervised training. The reason for including the second genetic stage is to iron out small changes that can improve the coloring quality. Letting the genetic training run for an extended period often leads to overfitting on the training dataset and, subsequently, poor performance on the test and validation datasets.

\section{Experiments}

\begin{figure*}[!ht]
\centering
\begin{tikzpicture}
\begin{axis} [xlabel=Epoch, ylabel=Loss, width=0.33\linewidth, at={(0\linewidth,0)}]
\addplot [red, mark=none] table [x=Epoch, y=sl-2] {training-results/loss.dat};
\addplot [blue, mark=none] table [x=Epoch, y=sl-3] {training-results/loss.dat};
\addplot [green, mark=none] table [x=Epoch, y=sl-4] {training-results/loss.dat};
\end{axis}

\begin{axis} [xlabel=Epoch, ylabel=Test F1-score, ymin=0.8, ymax=1.01, width=0.33\linewidth, at={(0.33\linewidth,0)},legend columns=3, legend style={at={(-0.12,-0.45)}, anchor=south west}]
\addplot [red, mark=none] table [x=Epoch, y=sl-2] {training-results/f1.dat};
\addplot [blue, mark=none] table [x=Epoch, y=sl-3] {training-results/f1.dat};
\addplot [green, mark=none] table [x=Epoch, y=sl-4] {training-results/f1.dat};
\end{axis}
\node[] at (7.77, 3.7) {\textbf{Smallest Degree Last}};

\begin{axis} [xlabel=Epoch, ylabel=Test colors, ymin=1650, ymax=1950, width=0.33\linewidth, at={(0.66\linewidth,0)}, domain=0:1000]
\addplot [red, mark=none] table [x=Epoch, y=sl-2] {training-results/colors.dat};
\addplot [blue, mark=none] table [x=Epoch, y=sl-3] {training-results/colors.dat};
\addplot [green, mark=none] table [x=Epoch, y=sl-4] {training-results/colors.dat};
\addplot [black, mark=none, dashed] {1888};
\node[] at (axis cs: 100,1868) {LF};
\addplot [black, mark=none, dashed] {1732};
\node[] at (axis cs: 100,1710) {SL};
\end{axis}

\begin{axis} [xlabel=Epoch, ylabel=Loss, ymin=0, width=0.33\linewidth, at={(0\linewidth,-158)}]
\addplot [red, mark=none] table [x=Epoch, y=sd-2] {training-results/loss.dat};
\addplot [blue, mark=none] table [x=Epoch, y=sd-3] {training-results/loss.dat};
\addplot [green, mark=none] table [x=Epoch, y=sd-4] {training-results/loss.dat};
\end{axis}

\begin{axis} [xlabel=Epoch, ylabel=Test F1-score, ymin=0.8, ymax=1.01, width=0.33\linewidth, at={(0.33\linewidth,-300)},legend columns=3, legend style={at={(-0.18,-0.55)}, anchor=south west}]
\addplot [red, mark=none] table [x=Epoch, y=sd-2] {training-results/f1.dat};
\addlegendentry{2-layer}
\addplot [blue, mark=none] table [x=Epoch, y=sd-3] {training-results/f1.dat};
\addlegendentry{3-layer}
\addplot [green, mark=none] table [x=Epoch, y=sd-4] {training-results/f1.dat};
\addlegendentry{4-layer}
\end{axis}
\node[] at (7.77, -1.2) {\textbf{Saturation Degree}};

\begin{axis} [xlabel=Epoch, ylabel=Test colors, ymin=1650, ymax=1950, width=0.33\linewidth, at={(0.66\linewidth,-425)}, domain=0:1000]
\addplot [red, mark=none] table [x=Epoch, y=sd-2] {training-results/colors.dat};
\addplot [blue, mark=none] table [x=Epoch, y=sd-3] {training-results/colors.dat};
\addplot [green, mark=none] table [x=Epoch, y=sd-4] {training-results/colors.dat};
\addplot [black, mark=none, dashed] {1888};
\node[] at (axis cs: 100,1868) {LF};
\addplot [black, mark=none, dashed] {1732};
\node[] at (axis cs: 100,1710) {SL};
\end{axis}

\end{tikzpicture}
\caption{Supervised training results with a varying number of layers. The top three plots show loss, F1-scores, and coloring quality for labels generated with the SL heuristic. Similarly, the bottom three plots shows the same for labels generated with the SD heuristic. The coloring plots also includes dashed lines indicating how many colors LF and SL used. }
\label{fig:supervised-training}
\end{figure*}
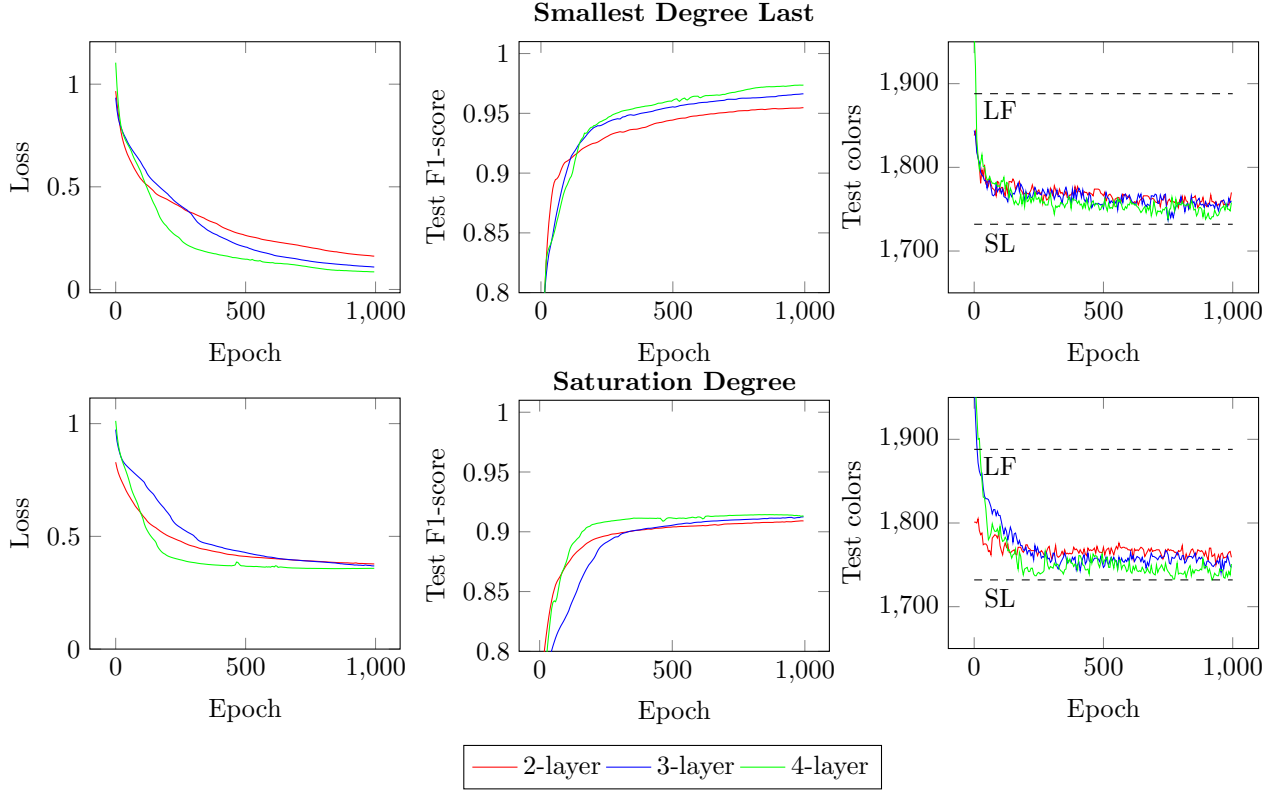

The following section first introduces the experimental setup and establishes the datasets used for training and evaluation. Then, a short outline of how the algorithms were implemented is provided. Finally, the training and experimental results are given.

\subsection{Instances}

The datasets used are comprised of 67 graphs from the Stanford large dataset collection SNAP \cite{snapnets} and 80 instances given as part of the second DIMACS implementation challenge for graph coloring \cite{johnson1996cliques}. Combined, these cover a wide range of graph types. The SNAP graphs include among other social networks, web graphs, and road networks. In contrast, the DIMACS instances cover various artificially generated graphs.

The graphs are divided into three datasets, and are ordered by increasing number of edges.
\begin{itemize}
    \item \textbf{Training.} The 40 smallest graphs from SNAP and the 60 smallest from DIMACS are used as the training dataset. In total, this includes over one million vertices.
    \item \textbf{Test.} The next 13 graphs from SNAP and 10 from DIMACS are used as the test dataset.
    \item \textbf{Validation.} The validation dataset consists of the remaining largest 14 graphs from SNAP and 10 from DIMACS. These instances start from around one million edges and include one very large instance with over one billion edges.
\end{itemize}

\subsection{Experimental Environment}

The training and experiments were run on a machine with an AMD EPYC 7302P 16-core processor and 128 gigabytes of memory, running Ubuntu 22.04.4 with Linux kernel 5.15.0-105. All algorithms were implemented in C and compiled with GCC version 11.4.0 using the -O3 flag. OpenMP is used for all the parallel implementations. The source code and trained models are openly available at GitHub\footnote{\url{https://github.com/KennethLangedal/GNN-COLORING}}. All experiments are repeated five times, and the best result is used.

\subsection{Implementation Techniques}

Due to the time-sensitive nature of greedy coloring heuristics and the goal of competing on both time and quality, it is important to eliminate as much execution overhead as possible. PyTorch Geometric is the most commonly used framework for implementing graph neural network models \cite{Fey/Lenssen/2019}. While PyTorch allows for easy development and prototyping, there will always be some inherent overhead due to the flexibility offered by the library.

A quick benchmark of PyTorch's GraphSAGE implementation on a few graphs from the validation dataset reveals that even a small three-layer model with 16 features falls far behind all but the most costly greedy heuristics, see Table \ref{tab:gnn-perf}. Besides PyTorch, and without much manual programming, the propagation from Algorithm \ref{alg:SAGE} can be implemented using optimized BLAS libraries such as OpenBLAS \cite{openblas}. The BLAS library only helps with the second transformation step, which is standard matrix multiplication, so it still requires a manual implementation of the message passing and activation function. Note that this approach is most likely similar to how PyTorch works internally. However, it still results in a significant improvement in terms of running time. Despite this, the OpenBLAS approach is still more than twice as slow as the SL heuristics.

The proposed implementation, referred to as \textit{C-GNN}, brings the execution time down sufficiently to compete with the other heuristics. The final execution time for the forward propagation can be seen in Table \ref{tab:gnn-perf}. The rest of this subsection describes the steps taken to achieve this execution time.

\begin{table}[ht]
    \centering
    \caption{Sequential execution time in seconds for the forward propagation of the GraphSAGE model with three layers. The implementations, highlighted in bold, are PyTorch geometric (PyT), OpenBLAS with manual message passing (OB), and the proposed implementation (C-GNN). For reference, the execution times for the greedy heuristics are also given. The number given for these greedy heuristics includes sequential ordering followed by greedy coloring. Results are given for three of the smallest graphs from the SNAP validation instances. }
    \vspace{10pt}
    \setlength{\tabcolsep}{3pt}
    \renewcommand{\arraystretch}{1.5}
    \scriptsize
    \begin{tabular}{l|lllllll}
        Instance    & FF    & \textbf{C-GNN}     & LF    & SL    & \textbf{OB}    & \textbf{PyT}   & SD    \\ \hline
        com-Youtube & 0.131 & \textbf{0.123} & 0.151 & 0.218 & \textbf{0.553} & \textbf{0.847} & 1.482 \\
        web-Google  & 0.163 & \textbf{0.162} & 0.218 & 0.391 & \textbf{0.764} & \textbf{1.176} & 1.739 \\
        wiki-Talk   & 0.269 & \textbf{0.238} & 0.276 & 0.381 & \textbf{0.961} & \textbf{1.521} & 4.173
    \end{tabular}
    \label{tab:gnn-perf}
\end{table}

\textbf{The number of hidden features is set to 16.} The reason for this number is solely hardware-related. One of the ways BLAS libraries achieve such high performance is by reducing data movement in and out of registers. This is achieved by accumulating results using a mostly square block of registers called a \textit{kernel}. A common shape is 2 by 4 registers \cite{goto2008anatomy}. On CPUs with 256-bit registers and single precision floats, this would be a kernel of size 16 by 4 floats. Therefore, having less than 16 features would come at the cost of lower CPU utilization. Going higher than 16 would result in a significant step-up in execution time, however, it would also increase the expressive power of the model.

\textbf{Graph storage format.} Graphs are stored in the compressed sparse row format (CSR). Since the edges are stored in continuous memory in CSR, some memory latency can be circumvented using software prefetching. The problem is that the next edge could point to any row in the $H^{(l)}$ matrix during the message passing stage. However, peeking ahead in the edge list shows exactly the row needed later, and a prefetch instruction can be executed in advance. This optimization also fits nicely with using 16 hidden features since each row of $H^{(l)}$ occupies exactly one cache line, assuming 64-byte cache lines. During initial testing, this improved the execution time by a factor between 1.2 to 1.5.

\textbf{Small number of parameters in each layer.} Following the previous two optimizations, the message-passing stage is currently memory-bound. Because each weighted transformation uses only a few parameters, it is possible to interleave the message passing and weighted transformation. The implementation works as follows: perform the message passing for a fixed number of vertices and store these rows in a buffer. When full, perform the matrix multiplication and store the corresponding rows of $H^{(l+1)}$. Assuming the parameters fit in the cache and that several memory reads are underway due to prefetching, this optimization can effectively hide large parts of the memory latency resulting from the irregular access during the message-passing step.

\textbf{No need to cache $H^{(l+1)}$.} Greedy coloring heuristics can color massive graphs very rapidly. At these sizes, there is little chance that by the time $H^{(l+1)}$ is needed at the next layer, it still resides in the cache. To utilize this observation, each write to $H^{(l+1)}$ can be done with a non-temporal memory write, removing the implicit read that would otherwise occur. This frees up more memory bandwidth to be used for reading the necessary rows of $H^{(l)}$.

For the other heuristics, we use techniques similar to those described in previous works \cite{hasenplaugh2014ordering}. Our implementation also performs on par with other implementations we have tested \cite{gebremedhin2013colpack}. Using the same implementation of greedy and JP for both the GNNs and the other heuristics is also beneficial, since we primarily want to evaluate the ordering part of the heuristics. Any improvement made to JP would benefit all the parallel heuristics equally. Sequentially, it is not always worthwhile to construct the ordering and then run greedy. For example, FF does not require any reordering of the vertices before coloring. Using software prefetching, sequential FF runs exceptionally fast, often more than four times faster than LF with greedy. We keep this sequential version of FF as a baseline.

\subsection{GNN Training Results}

For supervised training, models using two, three, and four layers are used, all with 16 hidden features. Since all the training graphs are fairly small, one mini-batch corresponds to one graph from the training dataset. The Adam optimizer~\cite{kingma2014adam} is used with its default hyperparameters and binary cross-entropy loss. Ground truth labels are generated using the SL or SD heuristics. Both are modified similarly to the SL formulation given by Allwright et al.~\cite{allwright1995comparison}. This means that every vertex fulfilling the minimum degree for SL or maximum saturation degree for SD during the construction of the ordering is assigned the same priority. The benefit of this is that the models need to learn less tiebreaking. The results from the training can be seen in Figure \ref{fig:supervised-training}. The training is also fast since we use shallow models with few hidden features. Running the training procedure sequentially for one model on the evaluation machine takes less than 30 minutes.

\begin{table*}[!t]
\setlength{\tabcolsep}{1.45pt}
\renewcommand{\arraystretch}{1.5}
\centering
\caption{Detailed results for the largest graphs from the validation dataset, presented in the same style as Hasenplaugh et al.~\cite{hasenplaugh2014ordering}. The $H$ column gives the name of the heuristic and $C$ the number of colors used. The number after GNN represents the number of layers in the model. $T_s$ is the sequential execution time. $T_1$ is the parallel implementation using a single thread, $T_{16}$ is the same for 16 threads without SMT, and $T_{32}$ is 32 threads with SMT. Speedup is shown in $T_1/T_{16}$ and $T_1/T_{32}$. Lastly, the slowdown when switching to a parallel implementation is shown in the $T_s/T_1$ column. }
    \vspace{10pt}
\scriptsize
\begin{tabular}{r|rrrrrrrrr|rrrrrrrrr}
Instance, $|V|$, $|E|$ & $H$  & $C$     & $T_s$     & $T_1$     & $T_s/T_1$ & $T_{16}$   & $T_{32}$   & $T_1/T_{16}$ & $T_1/T_{32}$ & $H$  & $C$     & $T_s$     & $T_1$     & $T_s/T_1$ & $T_{16}$   & $T_{32}$   & $T_1/T_{16}$ & $T_1/T_{32}$ \\ \hline
\textbf{twitter7}      & FF & 1,072 & 48.96  & 68.25  & 0.72  & 19.56 & 17.22 & 3.49   & 3.96   & GNN-2 & 1,103 & 122.93 & 124.55 & 0.99  & 19.50 & 16.20 & 6.39   & 7.69   \\
41,652,230    & LF & 1,077 & 72.71  & 73.69  & 0.99  & 13.99 & 12.18 & 5.27   & 6.05   & GNN-3 & 1,005 & 149.77 & 152.79 & 0.98  & 21.16 & 18.30 & 7.22   & 8.35   \\
2,405,026,092 & SL & 917   & 117.74 & 122.82 & 0.96  & 22.14 & 20.93 & 5.55   & 5.87   & GNN-4 & 1,031 & 181.72 & 183.93 & 0.99  & 23.93 & 20.47 & 7.69   & 8.98   \\ \hline
\textbf{com-Orkut}     & FF & 116   & 1.02   & 2.99   & 0.34  & 0.59  & 0.52  & 5.07   & 5.72   & GNN-2 & 85    & 7.02   & 7.17   & 0.98  & 0.76  & 0.59  & 9.40   & 12.19  \\
3,072,441     & LF & 87    & 4.50   & 4.72   & 0.95  & 0.55  & 0.39  & 8.57   & 12.14  & GNN-3 & 82    & 8.07   & 8.20   & 0.98  & 0.85  & 0.67  & 9.67   & 12.20  \\
234,370,166   & SL & 83    & 7.74   & 8.15   & 0.95  & 0.94  & 0.74  & 8.64   & 10.98  & GNN-4 & 82    & 9.38   & 9.51   & 0.99  & 0.97  & 0.79  & 9.80   & 12.05  \\ \hline
\textbf{LiveJournal1}  & FF & 325   & 0.47   & 1.81   & 0.26  & 0.33  & 0.28  & 5.54   & 6.46   & GNN-2 & 324   & 3.21   & 3.33   & 0.96  & 0.41  & 0.33  & 8.08   & 10.20  \\
4,847,571     & LF & 323   & 2.21   & 2.32   & 0.95  & 0.32  & 0.24  & 7.24   & 9.74   & GNN-3 & 322   & 3.66   & 3.79   & 0.97  & 0.45  & 0.36  & 8.50   & 10.60  \\
85,702,474    & SL & 322   & 3.42   & 3.67   & 0.93  & 0.50  & 0.39  & 7.33   & 9.32   & GNN-4 & 321   & 4.20   & 4.32   & 0.97  & 0.50  & 0.41  & 8.63   & 10.60  \\ \hline
\textbf{cit-Patents}   & FF & 15    & 0.26   & 1.28   & 0.20  & 0.11  & 0.08  & 11.42  & 15.91  & GNN-2 & 13    & 2.00   & 2.16   & 0.92  & 0.19  & 0.14  & 11.16  & 15.09  \\
3,774,768     & LF & 14    & 1.56   & 1.69   & 0.92  & 0.15  & 0.10  & 10.93  & 16.55  & GNN-3 & 13    & 2.31   & 2.50   & 0.92  & 0.23  & 0.18  & 10.79  & 14.27  \\
33,037,894    & SL & 13    & 2.26   & 2.50   & 0.90  & 0.23  & 0.16  & 10.79  & 15.52  & GNN-4 & 13    & 2.64   & 2.85   & 0.93  & 0.27  & 0.20  & 10.72  & 13.94  \\ \hline
\textbf{soc-Pokec}     & FF & 41    & 0.18   & 0.73   & 0.25  & 0.10  & 0.08  & 7.11   & 8.76   & GNN-2 & 33    & 1.41   & 1.48   & 0.95  & 0.15  & 0.12  & 9.77   & 12.75  \\
1,632,803     & LF & 34    & 0.89   & 0.96   & 0.93  & 0.11  & 0.08  & 8.55   & 12.73  & GNN-3 & 30    & 1.62   & 1.68   & 0.96  & 0.16  & 0.13  & 10.23  & 13.28  \\
44,603,928    & SL & 30    & 1.57   & 1.66   & 0.94  & 0.19  & 0.14  & 8.92   & 11.96  & GNN-4 & 33    & 1.92   & 1.97   & 0.97  & 0.19  & 0.15  & 10.51  & 13.33  \\ \hline
\textbf{wiki-topcats}  & FF & 61    & 0.15   & 0.53   & 0.27  & 0.38  & 0.41  & 1.39   & 1.29   & GNN-2 & 40    & 1.27   & 1.33   & 0.95  & 0.16  & 0.13  & 8.21   & 10.24  \\
1,791,489     & LF & 40    & 0.80   & 0.88   & 0.92  & 0.14  & 0.10  & 6.43   & 8.42   & GNN-3 & 40    & 1.44   & 1.52   & 0.95  & 0.17  & 0.14  & 8.86   & 10.95  \\
50,888,414    & SL & 41    & 1.42   & 1.55   & 0.92  & 0.21  & 0.18  & 7.28   & 8.77   & GNN-4 & 39    & 1.66   & 1.73   & 0.96  & 0.20  & 0.16  & 8.60   & 10.79  \\ \hline
\textbf{stackoverflow} & FF & 136   & 0.29   & 1.09   & 0.27  & 0.24  & 0.23  & 4.60   & 4.76   & GNN-2 & 88    & 1.92   & 1.96   & 0.98  & 0.28  & 0.25  & 6.88   & 7.76   \\
2,601,977     & LF & 88    & 1.30   & 1.34   & 0.97  & 0.24  & 0.21  & 5.51   & 6.42   & GNN-3 & 93    & 2.24   & 2.29   & 0.98  & 0.31  & 0.27  & 7.48   & 8.39   \\
56,367,036    & SL & 88    & 2.19   & 2.32   & 0.95  & 0.36  & 0.32  & 6.46   & 7.19   & GNN-4 & 90    & 2.64   & 2.69   & 0.98  & 0.34  & 0.30  & 7.99   & 8.96   \\ \hline
\textbf{wiki-Talk}     & FF & 80    & 0.04   & 0.27   & 0.17  & 0.10  & 0.11  & 2.75   & 2.52   & GNN-2 & 63    & 0.42   & 0.42   & 0.99  & 0.10  & 0.11  & 4.07   & 3.73   \\
2,394,385     & LF & 71    & 0.27   & 0.28   & 0.97  & 0.10  & 0.10  & 2.78   & 2.83   & GNN-3 & 59    & 0.50   & 0.51   & 0.99  & 0.11  & 0.12  & 4.53   & 4.09   \\
9,319,130     & SL & 57    & 0.36   & 0.38   & 0.94  & 0.12  & 0.12  & 3.28   & 3.18   & GNN-4 & 58    & 0.58   & 0.59   & 0.99  & 0.12  & 0.14  & 4.74   & 4.24   \\ \hline
\textbf{com-Youtube}   & FF & 39    & 0.03   & 0.13   & 0.20  & 0.03  & 0.03  & 4.20   & 4.50   & GNN-2 & 29    & 0.22   & 0.23   & 0.96  & 0.03  & 0.03  & 6.81   & 7.46   \\
1,134,890     & LF & 32    & 0.14   & 0.15   & 0.95  & 0.03  & 0.02  & 5.30   & 6.51   & GNN-3 & 28    & 0.26   & 0.27   & 0.96  & 0.04  & 0.03  & 7.30   & 8.24   \\
5,975,248     & SL & 29    & 0.21   & 0.22   & 0.92  & 0.04  & 0.04  & 5.52   & 6.23   & GNN-4 & 28    & 0.30   & 0.31   & 0.97  & 0.04  & 0.04  & 7.48   & 8.52  
\end{tabular}
\label{tab:large}
\end{table*}

Even the smallest 2-layer model can closely approximate the SL heuristics. Increasing to three and four layers improves the model's ability to learn the orderings, which is also reflected in the coloring quality. However, the data is noisy, and the difference between the models is small. All the models struggle more to learn the SD heuristic. Despite the reduced model accuracy, the coloring quality is slightly better than the models trained for the SL heuristic. While the trend indicates that more layers perform better, this is not expected to extend much further due to problems with oversmoothing caused when stacking too many GNN layers. Recent results are trying to address these issues \cite{kelesis2023reducing}, but it remains a challenge to train deep GNN networks.

These results do not include training using the LF or FF heuristics. This is because those heuristics are trivial to learn when both degree and ID are given as input features. Initial testing also confirmed this.

For the second stage with the genetic-based training, the top performing models in terms of colors used from the supervised stage are used as the initial population. In total, 100 models make up the first generation. Models with different numbers of layers are not mixed, and the whole genetic training stage is repeated for each model size. As mentioned earlier, the number of iterations is kept small to prevent overfitting to the training data. Therefore, we only expect the coloring quality to increase slightly. The results after 500 iterations can be seen in Table \ref{tab:genetic-training}.

\begin{table}[H]
    \centering
    \caption {Reduction in the number of colors used after 500 iterations of the genetic training. The values correspond to the number of colors used by the best model in the initial population minus the best in the final population. The entire genetic procedure was executed independently for each model size. }
    \vspace{10pt}
\begin{tabular}{l|ll}
        & Reduction Training & Reduction Test \\ \hline
2-layer & 164                & 10             \\
3-layer & 433                & 20             \\
4-layer & 430                & 53          

\end{tabular}
    \vspace{5pt}
    \label{tab:genetic-training}
\end{table}

\subsection{Experimental Results}

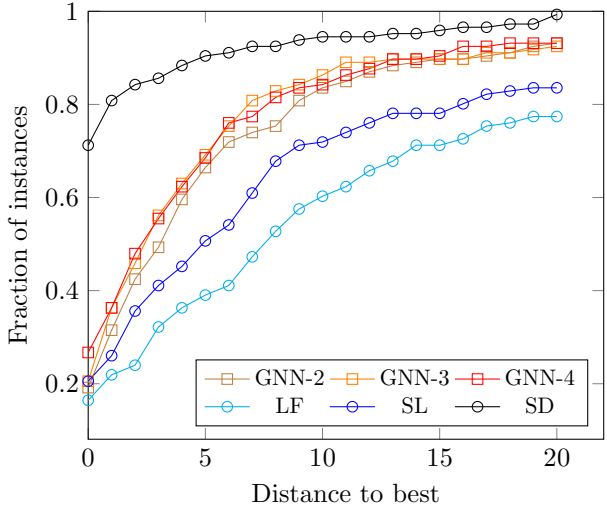
\begin{figure}[t]
\centering
\begin{tikzpicture}
\begin{axis}[
        xmin=0,
        ymax=1 ,
        legend columns=3,
        legend pos=south east,
        legend style={nodes={scale=0.85, transform shape}},
        ylabel={Fraction of instances},
        xlabel={Distance to best},
        ]
    \addplot[color=brown, mark=square] table[x=T, y=GNN-2] {data/PF.dat};
    \addplot[color=orange, mark=square] table[x=T, y=GNN-3] {data/PF.dat};
    \addplot[color=red, mark=square] table[x=T, y=GNN-4] {data/PF.dat};
    \addplot[color=cyan, mark=o] table[x=T, y=LF] {data/PF.dat};
    \addplot[color=blue, mark=o] table[x=T, y=SL] {data/PF.dat};
    \addplot[color=black, mark=o] table[x=T, y=SD] {data/PF.dat};
    \legend{GNN-2,GNN-3,GNN-4,LF,SL,SD,ID}
\end{axis}
\end{tikzpicture}
\caption{Performance profile for the entire dataset. The y-axis shows the fraction of instances solved using no more than $k$ additional colors compared to the best solution, where $k$ increases along the x-axis. At $k=0$, the value corresponds to the fraction of instances where the heuristic found the best solution. The number after GNN represents the number of layers in the model. }
\label{fig:pf-total}
\end{figure}

The proposed greedy GNN-based coloring heuristic is compared primarily to the LF and SL heuristics. In terms of state-of-the-art, most of the greedy heuristics mentioned so far are state-of-the-art on their own segment on the time axis. For example, FF would be state-of-the-art if you only care about the time aspect since nothing else will give you a feasible solution faster except assigning each vertex its own color. Similarly, all the other heuristics will typically give the best solution upon completion, assuming everything that takes longer is ignored. For this reason, we choose LF and SL as baselines to compare against since these are the best heuristics that are also fast and scalable. For reference, results for SD are also included in some figures. However, note that SD is often orders of magnitude slower and does not yet have an efficient parallel implementation.

Parallel scalability depends largely on the structure and size of the graph. Table \ref{tab:large} shows detailed results for the largest instances in the validation set. The GNN-based heuristics consistently give the best scalability. The average speedup using 16 threads is 8.46, compared to 7.09 with SL and 6.73 with LF. It also incurs little overhead when switching to a parallel implementation. On average, the parallel GNN with one thread is 3\% slower than its sequential version, compared to a 7\% slowdown with SL and 5\% with LF.

Using simultaneous multithreading (SMT) benefits all the parallel heuristics tested. The average speedup improves to 10.16 with the GNNs, 8.78 with SL, and 9.04 with LF. The difference is very noticeable in some cases, such as the cit-Patents graph. Only for the wiki-Talk graph does SMT seem to consistently hurt performance.

Notice that even for the largest instance in our dataset, twitter7, with over 1 billion edges, it took less than 20 seconds to compute colorings using the whole machine. For reference, it took 9 minutes for the ID heuristic using 1095 colors and over 2 hours for the SD heuristic using 917 colors.

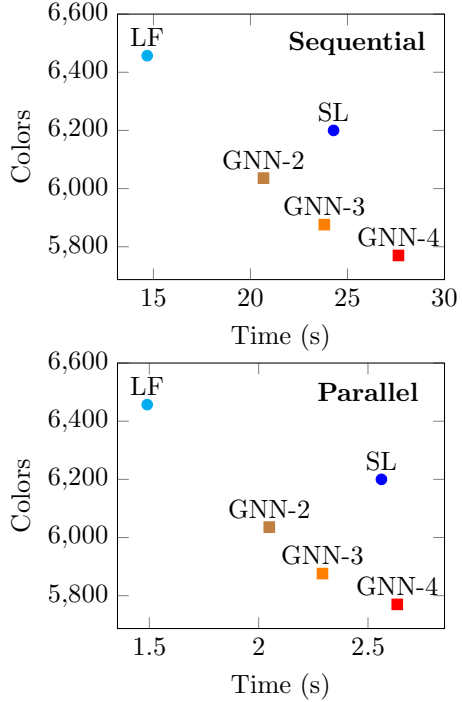
\begin{figure}[t]
\centering
\begin{tikzpicture} []

  \def\MarkSize{.75em}
  \protected\def\ToWest#1{%
    \llap{#1\kern\MarkSize}\phantom{#1}%
  }
  \protected\def\ToSouth#1{%
    \sbox0{#1}%
    \smash{%
      \rlap{%
        \kern-.5\dimexpr\wd0 + \MarkSize\relax
        \lower\dimexpr.375em+\ht0\relax\copy0 %
      }%
    }%
    \hphantom{#1}%
  }

  \begin{axis}[
    width=0.7\linewidth, at={(0,1200)},
    ylabel={Colors},
    xlabel={Time (s)},
    ymax=6600,
    xmax=30,
    scatter/classes={
            cyan={cyan},
            blue={blue},
            brown={mark=square*,brown},
            orange={mark=square*,orange},
            red={mark=square*,red}
        }
  ]
  \addplot[
    scatter,
    only marks,
    nodes near coords*={\data},
    visualization depends on={value \thisrow{label} \as \data},
    scatter src=explicit symbolic
  ] 
  table [x=x,y=y,meta=color]{data/SNAP-seq.dat};
  \node[] at (25.5, 6500) {\textbf{Sequential}};
  \end{axis}

  \begin{axis}[
    width=0.7\linewidth, at={(0,0)},
    ylabel={Colors},
    xlabel={Time (s)},
    ymax=6600,
    xmax=2.85,
    scatter/classes={
            cyan={cyan},
            blue={blue},
            brown={mark=square*,brown},
            orange={mark=square*,orange},
            red={mark=square*,red}
        }
  ]
  \addplot[
    scatter,
    only marks,
    nodes near coords*={\data},
    visualization depends on={value \thisrow{label} \as \data},
    scatter src=explicit symbolic
  ] 
  table [x=x,y=y,meta=color]{data/SNAP-par.dat};
  \node[] at (2.5, 6500) {\textbf{Parallel}};
  \end{axis}
\end{tikzpicture}
\caption{Accumulated results for all validation instances. The top figure shows results for sequential runs and the bottom for parallel ones. The parallel runs use parallel ordering and JP, using the whole machine. The x-y coordinates correspond to the total number of colors and time used by each algorithm. The number after GNN represents the number of layers in the model. Note that the number of colors used is the same regardless of whether the sequential or parallel version is used and that the axes are cropped and do not start from zero. }
\label{fig:validation-results}
\end{figure}

Figure \ref{fig:pf-total} shows the combined results for coloring quality on the entire dataset. The SD heuristic gives significantly better colorings on this dataset, as was also shown by Hasenplaugh et al.~\cite{hasenplaugh2014ordering}. Interestingly, this is not always the case, as Gebremedhin et al.~\cite{gebremedhin2013colpack} observed that SL outperformed SD on synthetic and scientific computation graphs. As mentioned, SD is still significantly slower and does not scale, so this result should not be interpreted as SD dominating the competition. Figure \ref{fig:validation-results} shows sequential and parallel execution time and the number of colors used for the validation dataset. As in the training procedure, adding more layers reduces the number of colors used, but at the cost of increased execution time. When running sequentially, the smallest 2- and 3-layer models outperform SL on execution time, using significantly fewer colors. Increasing to 4 layers gives slightly better colorings while running slower than SL. Using the whole machine (32 cores and SMT), the 4-layer model runs almost as fast as SL.

\subsection{Repeated Coloring}

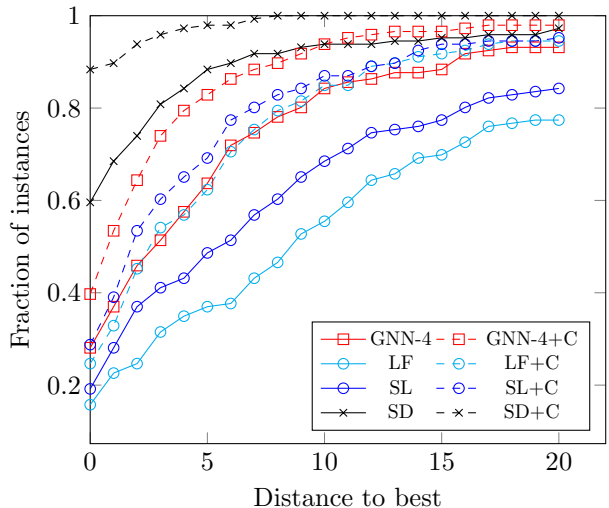
\begin{figure}[ht]
\centering
\begin{tikzpicture}
\begin{axis}[
        xmin=0,
        ymax=1 ,
        legend columns=2,
        legend pos=south east,
        legend style={nodes={scale=0.75, transform shape}},
        ylabel={Fraction of instances},
        xlabel={Distance to best},
        ]
    \addplot[color=red, mark=square] table[x=T, y=GNN4] {data/PFC.dat};
    \addplot[color=red, mark=square, dashed, mark options={solid}] table[x=T, y=GNN4+C] {data/PFC.dat};
    \addplot[color=cyan, mark=o] table[x=T, y=LF] {data/PFC.dat};
    \addplot[color=cyan, mark=o, dashed, mark options={solid}] table[x=T, y=LF+C] {data/PFC.dat};
    \addplot[color=blue, mark=o] table[x=T, y=SL] {data/PFC.dat};
    \addplot[color=blue, mark=o, dashed, mark options={solid}] table[x=T, y=SL+C] {data/PFC.dat};
    \addplot[color=black, mark=x] table[x=T, y=SD] {data/PFC.dat};
    \addplot[color=black, mark=x, dashed, mark options={solid}] table[x=T, y=SD+C] {data/PFC.dat};
    \legend{GNN-4,GNN-4+C,LF,LF+C,SL,SL+C,SD,SD+C}
\end{axis}
\end{tikzpicture}
\caption{Performance profile for the entire dataset including repeated application of Culberson. Each heuristic includes a \textit{C} version that represents the original coloring followed by five iterations of Culberson. The y-axis shows the fraction of instances solved using no more than $k$ additional colors compared to the best solution, where $k$ increases along the x-axis. The number after GNN represents the number of layers in the model. }
\label{fig:pfc-total}
\end{figure}

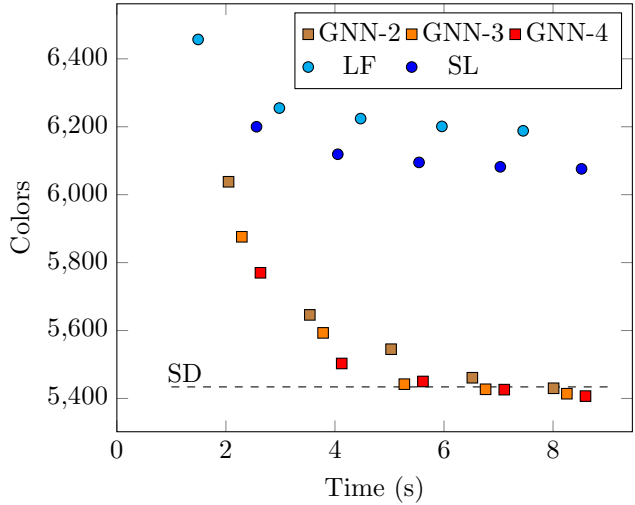
\begin{figure}[ht]
\centering
\begin{tikzpicture}
\begin{axis}[
        xmin=0,
        legend columns=3,
        ylabel={Colors},
        xlabel={Time (s)},
        scatter/classes={
            GNN-2={mark=square*, draw=black, fill=brown},
            GNN-3={mark=square*, draw=black, fill=orange},
            GNN-4={mark=square*, draw=black, fill=red},
            LF={mark=*, draw=black, fill=cyan},
            SL={mark=*, draw=black, fill=blue}
        }]
    \addplot[scatter, only marks, scatter src=explicit symbolic] table[meta=label] {data/SNAP-par-k.dat};
    \node [] at (1.25,5474) {SD};
    \draw [dashed] (1,5434) -- (9,5434);
    \legend{GNN-2,GNN-3,GNN-4,LF,SL}
\end{axis}
\end{tikzpicture}
\caption{Accumulated results for all validation instances using repeated application of Culberson with the parallel implementation. The x-y coordinates correspond to the total number of colors and time used. Subsequent marks of the same type, from left to right, correspond to repeated colorings. The number after GNN represents the number of layers in the model. Notice that the y-axis is cropped and does not start from zero.}
\label{fig:results-culberson}
\end{figure}

As a side note, Culberson introduced an iterative algorithm for graph coloring based on the observation that recoloring a graph where the vertices are processed according to their previous color class can never increase the number of colors used \cite{culberson1992iterated}. It can, however, often reduce the number of colors used. Figure \ref{fig:results-culberson} and \ref{fig:pfc-total} shows results for each heuristic using five repeated colorings, starting from the original assignment of colors and then repeatedly coloring the vertices ordered by their color from the previous coloring. In other words, repeated colorings set $P(u) = C(u), \forall u \in V$, where $C$ is the previous coloring. Even though every tested algorithm benefits from Culberson's approach, it is not enough to cause significant changes to our results. For example, the LF ordering with one iteration of Culberson uses almost the same number of colors as SL but is slightly slower. An interesting observation is that Culberson's repeated coloring effectively provides new coloring heuristics that use fewer colors. Furthermore, unlike the ID and SD heuristics, which do not have efficient parallel implementations, repeated colorings remain scalable. This is enough for our GNN-based heuristics to catch up to SD for the total number of colors used on the validation dataset. SD still performs better over the entire dataset, but the gap is significantly reduced.

\section{Conclusion and Future Work}

In this work, we have introduced a GNN-based greedy coloring heuristic that can compete with existing heuristics in coloring quality, execution time, and scalability. We used small models based on the GraphSAGE architecture with a tailor-made implementation to reach the speeds where existing heuristics operate. This result demonstrates for the first time that machine learning can outperform the fastest non-trivial coloring heuristics on both solution quality and execution time. Due to the low number of model parameters, supervised training is very inexpensive. Furthermore, the second part of the training using a genetic procedure could, in principle, improve the coloring quality for any dataset, even in cases where the other ordering heuristics might struggle. A brief summary of the results on coloring quality can be seen in Table \ref{tab:summary}.

As several others have mentioned, developing good parallel implementations for ID and SD remains an open problem \cite{hasenplaugh2014ordering}. On the machine-learning side, there are several other architectures and training techniques to try. As mentioned earlier, Kelesis et al.~\cite{kelesis2023reducing} have proposed techniques to address the oversmoothing issue caused by stacking too many GNN layers. To extend this work, one could use larger models and target the SD heuristic on quality.

\begin{table}[!h]
\centering
    \setlength{\tabcolsep}{4pt}
    \renewcommand{\arraystretch}{1.25}
\caption{Summary of colors used by each heuristic on each dataset. The number after GNN represents the number of layers in the model.}
    \vspace{10pt}
    \footnotesize
\begin{tabular}{l|llllll}
      & LF    & SL    & GNN-2 & GNN-3 & GNN-4 & SD    \\ \hline
Training & 5,789 & 5,310 & 5,334 & 5,268 & 5,252 & 4,942 \\
Test  & 1,888 & 1,732 & 1,722 & 1,702 & 1,649 & 1,508 \\
Validation   & 6,457 & 6,200 & 6,038 & 5,876 & 5,770 & 5,434
\end{tabular}
\label{tab:summary}
\end{table}

\printbibliography

@inproceedings{hasenplaugh2014ordering,
  title={Ordering heuristics for parallel graph coloring},
  author={Hasenplaugh, William and Kaler, Tim and Schardl, Tao B and Leiserson, Charles E},
  booktitle={Proceedings of the 26th ACM symposium on Parallelism in algorithms and architectures},
  pages={166--177},
  year={2014}
}

@techreport{culberson1992iterated,
    author={Culberson, Joseph},
    title={Iterated greedy graph coloring and the difficulty landscape},
    institution={University of Alberta},
    year={1992}
}

@inproceedings{Fey/Lenssen/2019,
  title={Fast Graph Representation Learning with {PyTorch Geometric}},
  author={Fey, Matthias and Lenssen, Jan E.},
  booktitle={ICLR Workshop on Representation Learning on Graphs and Manifolds},
  year={2019},
}

@article{kipf2016semi,
  title={Semi-supervised classification with graph convolutional networks},
  author={Kipf, Thomas N and Welling, Max},
  journal={arXiv preprint arXiv:1609.02907},
  year={2016}
}

@article{hamilton2017inductive,
  title={Inductive representation learning on large graphs},
  author={Hamilton, Will and Ying, Zhitao and Leskovec, Jure},
  journal={Advances in neural information processing systems},
  volume={30},
  year={2017}
}

@misc{PyG,
  title = {{PyTorch Geometric} Documentation},
  howpublished = {\url{https://pytorch-geometric.readthedocs.io/en/latest/index.html}},
  note = {Accessed: 2024-03-14}
}

@misc{openblas,
  title = {{OpenBLAS}},
  howpublished = {\url{http://www.openblas.net/}},
  note = {Accessed: 2024-03-14}
}

@article{schuetz2022graph,
  title={Graph coloring with physics-inspired graph neural networks},
  author={Schuetz, Martin JA and Brubaker, J Kyle and Zhu, Zhihuai and Katzgraber, Helmut G},
  journal={Physical Review Research},
  volume={4},
  number={4},
  pages={043131},
  year={2022},
  publisher={APS}
}

@article{such2017deep,
  title={Deep neuroevolution: Genetic algorithms are a competitive alternative for training deep neural networks for reinforcement learning},
  author={Such, Felipe Petroski and Madhavan, Vashisht and Conti, Edoardo and Lehman, Joel and Stanley, Kenneth O and Clune, Jeff},
  journal={arXiv preprint arXiv:1712.06567},
  year={2017}
}

@article{goto2008anatomy,
  title={Anatomy of high-performance matrix multiplication},
  author={Goto, Kazushige and Geijn, Robert A van de},
  journal={ACM Transactions on Mathematical Software (TOMS)},
  volume={34},
  number={3},
  pages={1--25},
  year={2008},
  publisher={ACM New York, NY, USA}
}

@book{karp2010reducibility,
  title={Reducibility among combinatorial problems},
  author={Karp, Richard M},
  year={2010},
  publisher={Springer}
}

@article{arkin1987scheduling,
  title={Scheduling jobs with fixed start and end times},
  author={Arkin, Esther M and Silverberg, Ellen B},
  journal={Discrete Applied Mathematics},
  volume={18},
  number={1},
  pages={1--8},
  year={1987},
  publisher={Elsevier}
}

@article{lewis2015timetable,
  title={A guide to graph colouring},
  author={Lewis, Rhyd},
  journal={Springer},
  volume={10},
  pages={195--218},
  year={2015},
  publisher={Springer}
}

@article{voloshin2009graph,
  title={Graph Coloring: History, results and open problems},
  author={Voloshin, V},
  journal={Alabama Journal of Mathematics, Spring/Fall},
  year={2009}
}

@article{barnier2004graph,
  title={Graph coloring for air traffic flow management},
  author={Barnier, Nicolas and Brisset, Pascal},
  journal={Annals of operations research},
  volume={130},
  pages={163--178},
  year={2004},
  publisher={Springer}
}

@article{welsh1967upper,
  title={An upper bound for the chromatic number of a graph and its application to timetabling problems},
  author={Welsh, Dominic JA and Powell, Martin B},
  journal={The Computer Journal},
  volume={10},
  number={1},
  pages={85--86},
  year={1967},
  publisher={Oxford University Press}
}

@article{lovasz1989line,
  title={An on-line graph coloring algorithm with sublinear performance ratio},
  author={Lov{\'a}sz, L{\'a}szl{\'o} and Saks, Michael and Trotter, William T},
  journal={Discrete Mathematics},
  volume={75},
  number={1-3},
  pages={319--325},
  year={1989},
  publisher={Elsevier}
}

@article{matula1983smallest,
  title={Smallest-last ordering and clustering and graph coloring algorithms},
  author={Matula, David W and Beck, Leland L},
  journal={Journal of the ACM (JACM)},
  volume={30},
  number={3},
  pages={417--427},
  year={1983},
  publisher={ACM New York, NY, USA}
}

@article{brelaz1979new,
  title={New methods to color the vertices of a graph},
  author={Br{\'e}laz, Daniel},
  journal={Communications of the ACM},
  volume={22},
  number={4},
  pages={251--256},
  year={1979},
  publisher={ACM New York, NY, USA}
}

@article{allwright1995comparison,
  title={A comparison of parallel graph coloring algorithms},
  author={Allwright, JR and Bordawekar, R and Coddington, PD and Dincer, K and Martin, CL},
  journal={SCCS-666},
  pages={1--19},
  year={1995},
  publisher={Citeseer}
}

@inproceedings{besta2020high,
  title={High-performance parallel graph coloring with strong guarantees on work, depth, and quality},
  author={Besta, Maciej and Carigiet, Armon and Janda, Kacper and Vonarburg-Shmaria, Zur and Gianinazzi, Lukas and Hoefler, Torsten},
  booktitle={SC20: International Conference for High Performance Computing, Networking, Storage and Analysis},
  pages={1--17},
  year={2020},
  organization={IEEE}
}

@article{newsam1983estimation,
  title={Estimation of sparse Jacobian matrices},
  author={Newsam, Garry N and Ramsdell, John D},
  journal={SIAM Journal on Algebraic Discrete Methods},
  volume={4},
  number={3},
  pages={404--418},
  year={1983},
  publisher={SIAM}
}

@article{jones1993parallel,
  title={A parallel graph coloring heuristic},
  author={Jones, Mark T and Plassmann, Paul E},
  journal={SIAM Journal on Scientific Computing},
  volume={14},
  number={3},
  pages={654--669},
  year={1993},
  publisher={SIAM}
}

@inproceedings{ijaz2022solving,
  title={Solving graph coloring problem via graph neural network (gnn)},
  author={Ijaz, Ali Zeeshan and Ali, Raja Hashim and Ali, Nisar and Laique, Talha and Khan, Talha Ali},
  booktitle={2022 17th International Conference on Emerging Technologies (ICET)},
  pages={178--183},
  year={2022},
  organization={IEEE}
}

@inproceedings{lemos2019graph,
  title={Graph colouring meets deep learning: Effective graph neural network models for combinatorial problems},
  author={Lemos, Henrique and Prates, Marcelo and Avelar, Pedro and Lamb, Luis},
  booktitle={2019 IEEE 31st International Conference on Tools with Artificial Intelligence (ICTAI)},
  pages={879--885},
  year={2019},
  organization={IEEE}
}

@article{galinier2006survey,
  title={A survey of local search methods for graph coloring},
  author={Galinier, Philippe and Hertz, Alain},
  journal={Computers \& Operations Research},
  volume={33},
  number={9},
  pages={2547--2562},
  year={2006},
  publisher={Elsevier}
}

@article{faycal2022direct,
  title={Direct mutation and crossover in genetic algorithms applied to reinforcement learning tasks},
  author={Faycal, Tarek and Zito, Claudio},
  journal={arXiv preprint arXiv:2201.04815},
  year={2022}
}

@inproceedings{montana1989training,
  title={Training feedforward neural networks using genetic algorithms.},
  author={Montana, David J and Davis, Lawrence and others},
  booktitle={IJCAI},
  volume={89},
  pages={762--767},
  year={1989}
}

@misc{snapnets,
  author       = {Jure Leskovec and Andrej Krevl},
  title        = {{SNAP Datasets}: {Stanford} Large Network Dataset Collection},
  howpublished = {\url{http://snap.stanford.edu/data}},
  month        = jun,
  year         = 2014
}

@book{johnson1996cliques,
  title={Cliques, coloring, and satisfiability: second DIMACS implementation challenge, October 11-13, 1993},
  author={Johnson, David S and Trick, Michael A},
  volume={26},
  year={1996},
  publisher={American Mathematical Soc.}
}

@article{kingma2014adam,
  title={Adam: A method for stochastic optimization},
  author={Kingma, Diederik P and Ba, Jimmy},
  journal={arXiv preprint arXiv:1412.6980},
  year={2014}
}

@incollection{kelesis2023reducing,
  title={Reducing Oversmoothing in Graph Neural Networks by Changing the Activation Function},
  author={Kelesis, Dimitrios and Vogiatzis, Dimitrios and Katsimpras, Georgios and Fotakis, Dimitris and Paliouras, Georgios},
  booktitle={ECAI 2023},
  pages={1231--1238},
  year={2023},
  publisher={IOS Press}
}

@article{gebremedhin2013colpack,
  title={Colpack: Software for graph coloring and related problems in scientific computing},
  author={Gebremedhin, Assefaw H and Nguyen, Duc and Patwary, Md Mostofa Ali and Pothen, Alex},
  journal={ACM Transactions on Mathematical Software (TOMS)},
  volume={40},
  number={1},
  pages={1--31},
  year={2013},
  publisher={ACM New York, NY, USA}
}

@article{li2022rethinking,
  title={Rethinking Graph Neural Networks for the Graph Coloring Problem},
  author={Li, Wei and Li, Ruxuan and Ma, Yuzhe and Chan, Siu On and Pan, David and Yu, Bei},
  journal={arXiv preprint arXiv:2208.06975},
  year={2022}
}

@article{huang2019coloring,
  title={Coloring big graphs with alphagozero},
  author={Huang, Jiayi and Patwary, Mostofa and Diamos, Gregory},
  journal={arXiv preprint arXiv:1902.10162},
  year={2019}
}

\end{document}